# Semantic Sensor Network Ontology based Decision Support System for Forest Fire Management

Ritesh Chandra[1], Kumar Abhishek[2], Sonali Agarwal[1] and Navjot Singh[1]

*Indian Institute of Information Technology Allahabad[1], National Institute of Technology Patna[2], India.*

*rsi2022001@iiita.ac.in, kumar.abhishek.cse@nitp.ac.in, sonali@iiita.ac.in, navjot@iiita.ac.in*

## Abstract

The forests are significant assets for every country. When it gets destroyed, it may negatively impact the environment, and forest fire is one of the primary causes. Fire weather indices are widely used to measure fire danger and are used to issue bushfire warnings. It can also be used to predict the demand for emergency management resources. Sensor networks have grown in popularity in data collection and processing capabilities for a variety of applications in industries such as medical, environmental monitoring, home automation etc. Semantic sensor networks can collect various climatic circumstances like wind speed, temperature, and relative humidity. However, estimating fire weather indices is challenging due to the various issues involved in processing the data streams generated by the sensors. Hence, the importance of forest fire detection has increased day by day. The underlying Semantic Sensor Network (SSN) ontologies are built to allow developers to create rules for calculating fire weather indices and also the convert dataset into Resource Description Framework (RDF). This research describes the various steps involved in developing rules for calculating fire weather indices. Besides, this work presents a Web-based mapping interface to help users visualize the changes in fire weather indices over time. With the help of the inference rule, it designed a decision support system using the SSN ontology and query on it through SPARQL. The proposed fire management system acts according to the situation, supports reasoning and the general semantics of the open-world followed by all the ontologies.

## I. INTRODUCTION

Forest fire is one of the major causes of the large-scale destruction of forest ecosystems. Considering the example of India, most of the fires are ignited due to anthropogenic activities, and there is limited evidence of fires due to natural causes such as lightning [27]. As per Sendai framework [26] priorities, there is a need to shift disaster monitoring to disaster risk assessment. Fire danger rating systems are practical tools for fire officials to manage fire controlling activities and can help to achieve the Sendai framework initiatives. Forest fire alerts are disseminated and uploaded into the Bhavan and Forest Survey of India websites [1] during the fire season to manage forest fire activities. Nevertheless, there is a limitation of fire alerts, such as fires that are actively burning at the time of satellite overpass can only be detected, whereas fires between the passes are likely to be missed. So, there is a need for a comprehensive and proactive system to forecast the possibility and severity of forest fires. There is limited availability of systems to detect forest fires and related hazards. Many forest fire cases are also evident across the globe. From 1980 to 2005, over 2.7 million hectares of forest in Portugal were destroyed by wildfires [39]. More than 24 million hectares (59 million acres) burned during Australia's devastating "Black Summer" bushfire season of 2019-2020. Global fire danger rating systems such as Canadian Forest Fire Danger Rating System, McArthur Forest Fire Danger Index [37], and US National Fire Danger Rating System [38] provide a global framework but are not very accurate at local, regional levels. These systems require large datasets consisting of a different period, daily automatic weather stations data, and ground investigation data.

Ontology is a conceptual framework that sets the relationship among various concepts related to a given domain. It provides a framework for carrying out tasks related to knowledge exchange. As a result, the basic concept of the semantic web has turned to a higher level, and different kinds of ontologies have been designed. The SSN ontology can portray sensors as abilities, organizations, observations, measurements that discuss various ontology concepts, ontology mapping design, building environments, algorithms etc. Following this the first step in developing a calculated area portrayal is utilizing a significant cosmology level to frame the structure's skeleton. Though there are numerous upper-level ontologies; the present research focuses on SSN Ontology [6].

A Decision Support System (DSS) is a computer-based information system that integrates models and data to handle unstructured or semi-structured problems with multiple user engagement via a friendly user interface. The SSN Ontology-based DSS for forest fire management is proposed here to prevent different types and stages of fires that help to save our natural resources, life, global warming, etc. The proposed system is specifically built to meet the following goals and user requirements:

- To improve the quality of data, we convert raw data into Resource Description Framework (RDF) triples.
- This research work aims to create an index that displays fire weather threat ratings based on meteorological expert input. (In the form of Web Ontology Language (OWL) [36]).
- To design a Semantic Web Rule Language (SWRL) for a rule-based DSS system which can be expressed as classes, properties and individuals.
- To create a collection of online services that facilitates the fire weather conditions in a specific location to be searched, explored, and visualized by the users.

The remaining part of the paper is organized in various sections in which literature review and similar initiatives are discussed in section 2. Section 3 provides the details of the Montesinho natural park dataset. Section 4 introduces the SSN ontology and related details. In section 5, the RDF data conversion has been elaborated. and detailed information about the evaluation process of the fire weather index and SWRL rules are presented. Section 6 discusses the implementation setup and SPARQL query results are explained in section 7. The concluding remarks and future directions are summarized in section 8.

## II. RELATED WORKS

Telemetered data collected by satellite is the main method used to forecast and detect forest fires. However, it is not always possible to combine satellite images with other data sources to detect fire spots and improve wildfire monitoring [1], [2]. The FireWatch system was developed to improve the detection of forest fires using WSNs and other satellite-based technologies. It was also designed to support fire hazard predictions [3].
Sensor-based fire detection systems have been introduced to offer suppression for the monitoring system in the early stages of detection. Among them are temperature sensors, smoke sensors, infrared sensors, optical sensors, gas sensors, and various types of fire alarm sensors [12]–[15]. Using a laser spectroscopic carbon monoxide sensor, a fire detection system was proposed. Used a basic digital lock-in amplifier (DLIA), and a highly effective microcontroller for early fire detection [11]. A long-range Raman distributed fiber temperature sensor (RDFTS) was developed for early fire detection, with a maximum sensing distance of 30 kilometers and a spatial resolution of 28 meters. A temperature sensor can be used to measure temperature at a specific location, but it is not ideal for long-range sensing. Wang and Wang [17] presented a photoacoustic gas sensor based on wavelength modulation spectroscopy and using a near-infrared tunable fiber laser. Under atmospheric pressure, this sensor offers quick and concentrated readings of combustion products, particularly $C_2$, $H_2$ and $CO_2$. On the other hand, chemical gas sensors are more responsive than smoke particle detectors. While a sensor-based fire detection system, due to the requirement of frequently distributed sensors in close proximity, is not feasible.
Jerome V et al. [18] in this study, it offer an autonomous approach for detecting early smoke sources using landscape photos in real time. The first section discusses the segmentation technique it apply to extract pixels' persistent dynamical envelopes. Then, based on an analysis of the velocity of smoke plumes, it give our main criterion for smoke recognition.
Emmy Premal C et al. [19] proposed categorizing pixel flames using a YCbCr color scheme and statistical parameters such as mean and standard deviation derived for each Y, Cb, and Cr of the fire image. YCbCr images are more resistant to switching from luminance to chrominance than RGB ones. Two rules are used to segment the fire zone's central component, while the other two are used to segment the fire region. Finally, the product of this well-thought-out project achieves a fire detection rate of 99.4 %.
Weir, J. K. [16] traditional fire control, according to indigenous peoples, can supplement existing wildfire mitigation measures. There are many obstacles to adopting Indigenous methods into current wildfire legislation. The resurrection of Indigenous fire management in countries such as Australia is contingent, relying on ordinary persuasive labor and precarious intercultural diplomacy.

## III. STUDY AREA

This work aims to study the effects of forest fires on the Montesinho natural park [29]. The data collected during this period is taken from January 2000 to December 2003.The average annual temperature within the park is within the range of 8 to 12 degrees Celsius.The first database[2] is created by the Polytechnic Institute of Montesco. It contains data on the total burned area, the type of vegetation that affected the area, and the weather observations taken by a meteorological station in the park. As the vegetation type presented a low quality, only the X and Y axes are included. The Montesinho fire inspector's data is also taken into account. The weather conditions during a given month can affect the frequency of forest fires since human-caused ones can start. When a fire is discovered, the data collected by the station sensors are also used to alert firefighters.

[2]http://www.dsi.uminho.pt/ pcortez/forest fires/.

## IV. THE SEMANTIC SENSOR NETWORK ONTOLOGY

The Semantic Sensor Network (SSN) ontology was developed by the Semantic Sensor Networks Incubator Group [4] belonging to the World Wide Web Consortium (W3C). The semantic ontology can model various systems and sensor devices and their environment. The SSN is an ontology that enables developers to model and analyze sensor networks. When the environment changes, a sensor produces a stimulus. The object of the stimulus is then turned into another stimulus after it is detected. The properties of the sensors are observable in order to obtain meaningful results [5]. Several types of sensors, such as smoke sensors and wind sensors, can be placed in various location for forest fire detection.

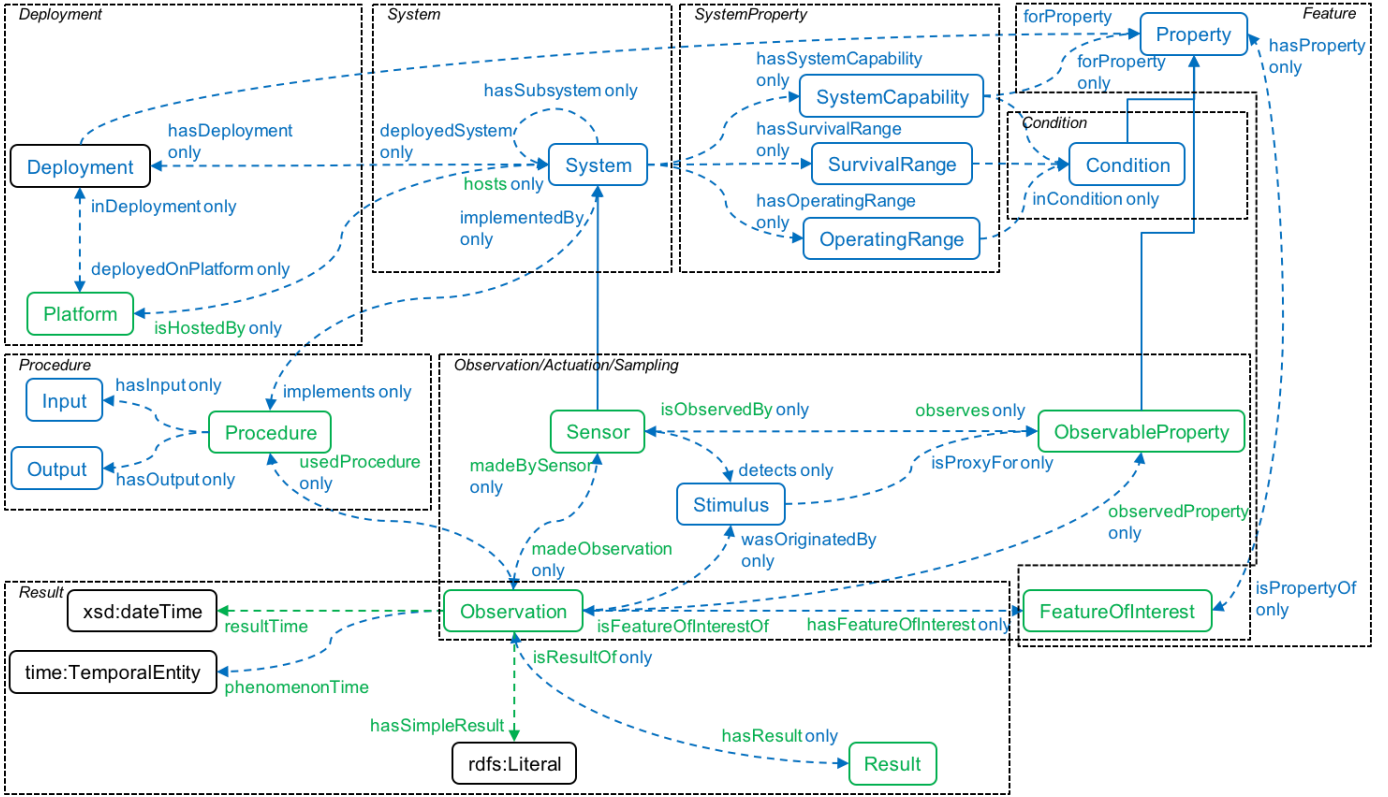

Fig. 1. **Core Module of SSN Ontology.** Which show the core classes and data properties, developed by W3C group [35].The subclass relationships between the SSNO classes are represented by the dashed arrows. Dotted boxes are core classes.

The majority of the ontology is thus an expansion of these classes, with subclasses and new attributes, as well as notions relevant to forest fire properties. Using OWL constructs, the SSN forest fire ontology models the ideas, relationships, attributes, datatypes, and limitations of the water forest fire detection. Observations are the connections between a sensor and its output. It can be used to identify interesting features in the environment. According to the sensing technique, the features must be fixed. A sensing method can also be used to describe the events that happened in real-time, for instance, how a sensor was positioned or used. The SSN has established the basic concepts and relationships (features, observations, characteristics, observations, systems, and sensors) and has aligned them with the DOLCE-Ultra Lite (DUL) ontology [7].

## V. RESOURCE DESCRIPTION FRAMEWORK (RDF) CONVERSION

The Resource Description Framework is a conceptual framework that describes the relationships between data types and their corresponding data structures. It supports the evolution of data structures without requiring the changes of data consumers. The RDF data model is a conceptual framework that takes advantage of the concept of triples to make statements about resources. It bases its calculations on a subject-predicate relationship (subject–predicate–object) [20]. As shown in Fig.2.

The RDF database is mapped to the SSN ontology in the second step of the process [8], [9]. The technique of finding correspondences between two sets of discrete items is known as ontology mapping. This method can be used to discover relationships between entities linked to a particular source ontology. One of the factors that should be considered when exploring the possibility of obtaining precise correspondences between various entities is to find a similarity measure that can allow us to perform the task. If we have a sensor table with the following attributes: ***sensor_location, observation_value, and***

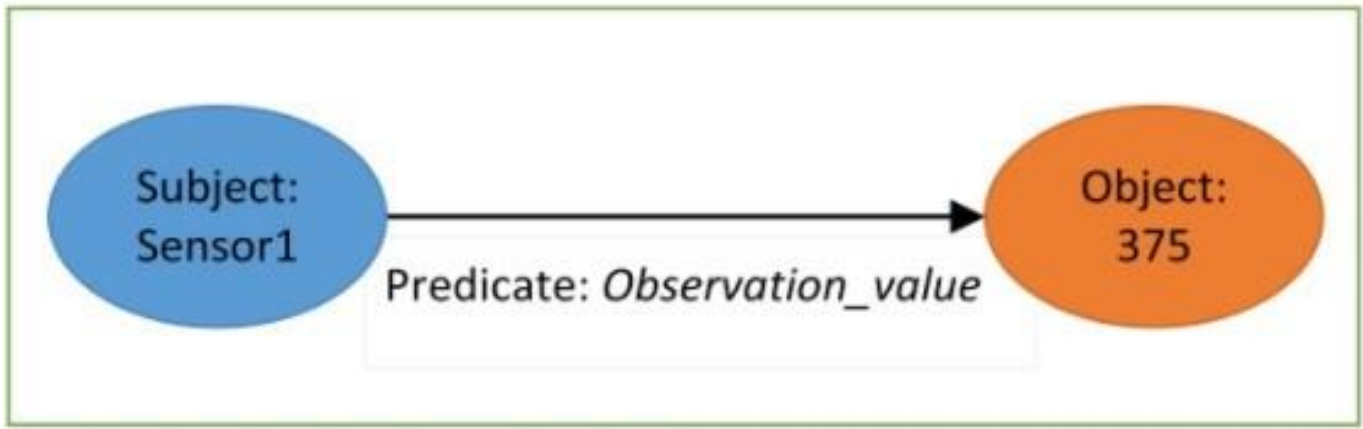

Fig. 2. **Example of RDF triple**

*observation_time* then we can obtain an RDF triple. For instance, in Figure 2 the triple represents the observed value(375) of a sensor. Some of the correspondence between the RDF sensor database elements and the SSN ontology are presented in Table 1.

Table 1. **The elements in the RDF sensor database correspond to the classes and characteristics in SSN.**

| RDF Database | SSN element | Description |
|---|---|---|
| Observation value | ssn:class:SensorOutput | Output of the sensor |
| has Output | ssn:property:isProducedBy | The output of the sensor |
| has observation value | ssn:property:hasvalue | Sensor value |
| measure unit | ssn:class:UnitofMeasure | Unit of observation data |
| sensor | ssn:class:sensing device | Sensor dentifier |
| has location | ssn:property:hasDeplyoment | The sensor location |
| has measure | ssn:property:Observes | Type of measure of sensor |

## VI. ONTOLOGY OF THE FIRE WEATHER INDEX

This research work proposes a Fire Weather Index (FWI) ontology to define fire weather index classes and associated features. It can take the ranges of Fuel Moisture Codes (FFMC, DMC, DC) and Fire Behaviour index (ISI, BUI, FWI) [28]. FFMC shows the ignition potential of the fire material, DMC and DC show the mop-up needs. ISI shows the rate of spread of fire, BUI shows the difficulty of control, and FWI shows the fire intensity which in shown in Figure 4.

In Fig.3, sensors give the weather observation like rain, temperature, wind, relative humidity through algorithms to convert into fuel moisture codes (follow the Canadian approach). DC Combines of Rain and Temperature. DMC combines rain, relative humidity, and temperature. FFMC is the combination of rain, relative humidity, temperature, and wind [23]. ISI combines FFMC and wind. BUI combines DMC and DC. FWI combines ISI and BUI. The result of FWI changes according to the sensor output. For Checking all the situations, we make a model checker, which is shown below Fig.4.

For calculating FMC it need water mass(Wf) and fuel mass(Wd). The formula for FMC is taken from Candian forest fire Management [24].

$$FMC = [(Wf − Wd)/Wd] × 100 \qquad (1)$$

Using the Fine Fuel Moisture Code (FFMC), it calculated the FMC's expected values using the following formula:

$$FMC = 147.2 × (101 − FFMC)/(59.5 + FFMC) \qquad (2)$$

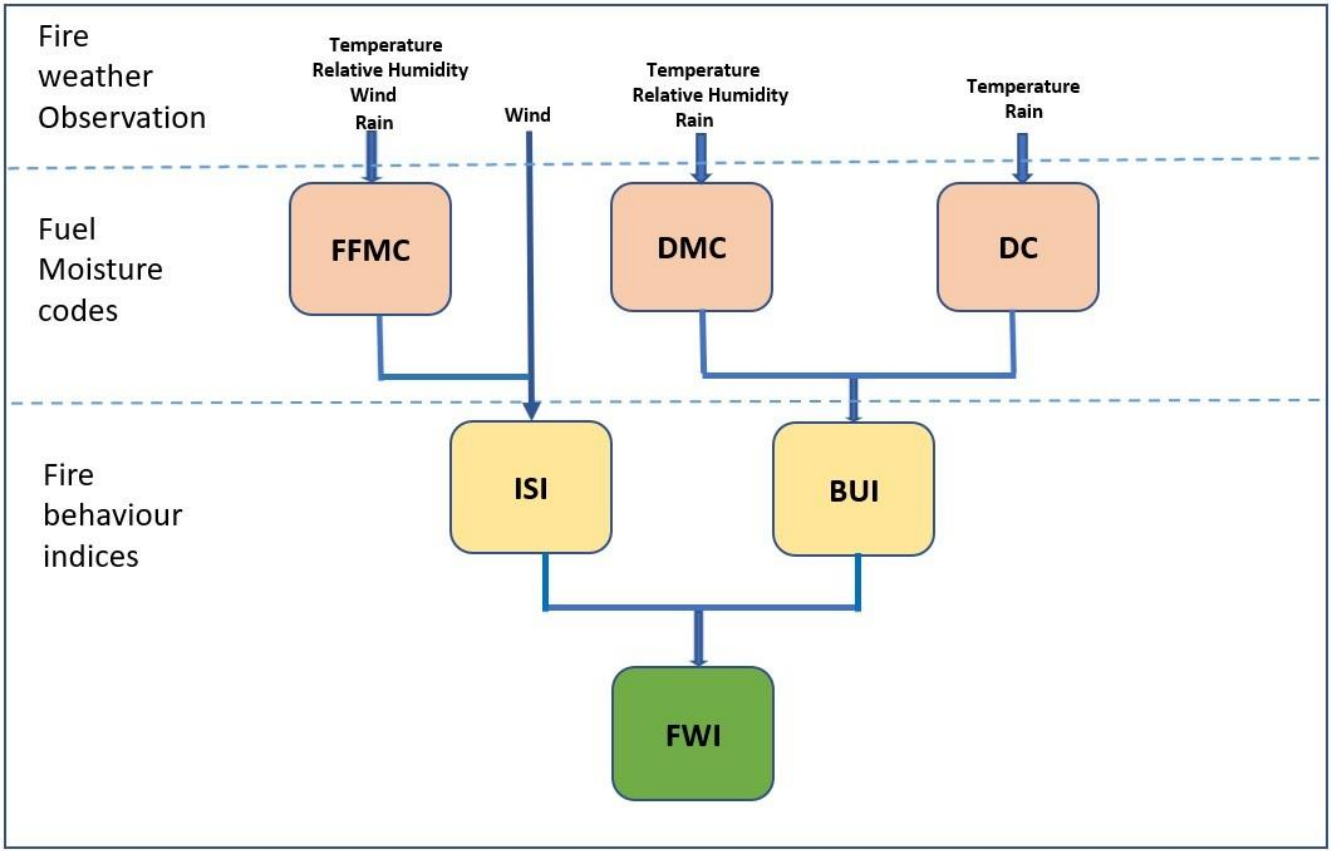

Fig. 3. **Fire Weather Index Flowchart**

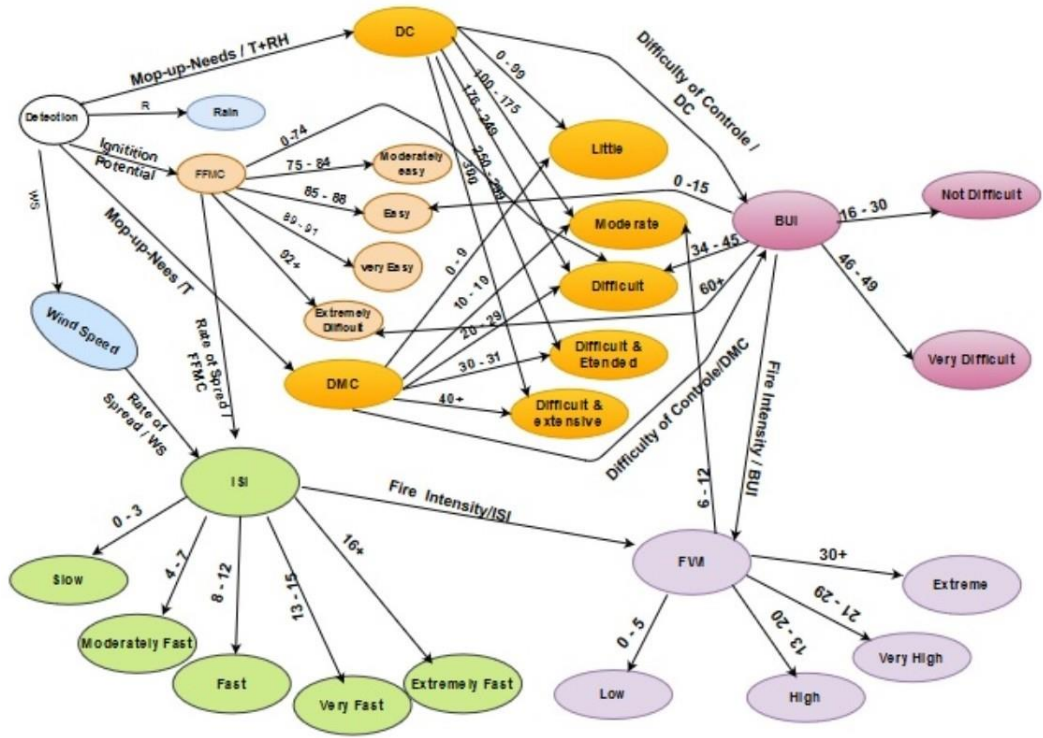

Fig. 4. **Model checker for Forest Fire**

In Fig. 4 represents Model checker for Forest fire which gives the complete precautionary measure detection according to the situation. The ranges included in this Model Checker from Canadian Forest Management System [28]. This ranges categorised in different precautionary measure situation (i.e. low,high,difficult etc). For more clarification following through this **Example** if fire intensity extremes in any location of forest, it first check the ignition potential of nearest location available and compare with FFMC Ranges (as shown in Model Checker ). If ranges comes difficulty, then go for the difficulty of control and compare the ranges of BUI. If the difficulty is more severe according to the given range, then fire management takes action accordingly. Same way all other Precautionary measure taken place. For more accurately result performance this model Converted into SSN Ontology which have additional feature to query on it, also have complete taxonomy for better understanding which are discussed in further section.

---

**Algorithm 1** Algorithm for FWI

---

**Require:** Input : T1:value; T2 : Rain; T3: windspeed; T4 for checking FFMC, DC, DMC.

     Check t4;

 **For**

 FFMC: ignition potential

 Check T1                                        ▷ comparison of its ranges.

 Check T2                                        ▷ raining or not.

 **Goto**

 ISI: Fire intensity.

 Check T1                                        ▷ comparison of its ranges.

 **Goto**

 FWI : fire weather index.                             ▷ otput.

 **Else**

 DMC and dc : MoupNeeds

 Check T1                                        ▷ comparison of its ranges.

 Check T2                                        ▷ raining or not.

 **Goto**

 BUI : Difficulty of controle.

 check T1                                        ▷ raining or not.

  **Goto**

 FWI:fire weather index.                                ▷ output.

---

## VII. DEVELOPING SWRL RULES

The Web Rule Language or SWRL[3] is a combination of the Web Ontology Language and the Rule Markup Language. It provides a high level abstract structure for Horn-like rules [10]. All rules are expressed in terms of OWL concepts (classes, properties, individuals). Total 42 rules are included in this model.

**Fine Fuels Moisture Code (FFMC)** - is a numerical rating that shows the combustibility and start effectiveness of fine fills (i.e needles, surface litter, small twigs, leaves etc). It derived from yesterday FFMC data and noon time temperature, wind speed, relative humidity, and 24 hour rainfall. Ranges belongs 0 to 100 and it sub divided into five groups. First group 0 to 74, second group 75 to 84, third group 85 to 88, fourth group 89 to 92 and fifth gorup 92 to 100.. According to this ranges "Ignition Potential" are observe ( i.e difficult, moderately easy, very easy etc.). Based on this ranges the Inference rules are

---

made for give restriction on the ontology which shown in Table 2.

Table 2. **Fine Fuels Moisture Code swrl rules**

| Rules Created for FFMC | Explanation |
|---|---|
| sensor_id(?s)∧Difficult(?s,?rh)∧ swrl:greaterThan(?rh,"1")⇒ IgnitionPotential(?s, difficult) | If the ignition potential of fine fuels is greater than "1", then control of fire is "difficult". |
| sensor_id(?s)∧ Moderatelyeasy(?s,?rh)∧swrl:greaterThan(?rh,"75")⇒ IgnitionPotential(?s, moderatelyeasy) | If the ignition potential of fine fuels is greater than "75", then controlling fire is "moderately easy". |
| sensor_id(?s)∧(?s,?rh)∧swrlb:greaterThan(?rh,"85") ⇒IgnitionPotential(?s,easy) | If the ignition potential of fine fuels is greater than "85", then controlling fire is "easy". |
| sensor_id(?s)∧veryeasy(?s,?rh)∧swrl:greaterThan(?rh,"89") ⇒ IgnitionPotential(?s, veryeasy) | If the ignition potential of fine fuels is greater than "89", then controlling fire is "very easy". |
| sensor_id(?s)∧extremelyeasy(?s,?rh)∧swrl:greaterThan (?rh,"92") ⇒IgnitionPotential(?s, extremelyeasy) | If the ignition potential of fine fuels is greater than "92", then controlling fire is "extremely easy". |

**Duff Moisture Code (DMC)** - The sogginess rating is a numerical code that shows the significance of fire to a particular layer of thick material. This code is usually used to describe the substance's texture. The layer of partially and completely degraded plant substances that underlies the litter and directly just above soil surface Depth varies between 2 and 10 cm. It derived from yesterday DMC data and noon time temperature, wind speed, relative humidity, and 24 hour rainfall. Ranges belongs 0 to unlimited but it sub divided into five groups. First group 2 to 9, second group 10 to 19, third group 21 to 29, fourth group 30 to 39 and fifth group 92 to unlimited. According to this ranges "Mop-up-Needs" are observe ( i.e little, moderately , difficult extensive etc.). Based on this ranges the Inference rules are made for give restriction on the ontology which shown in Table 3.

Table 3. **Duff Moisture Code swrl rules**

| Rules Created for DMC | Explanation |
|---|---|
| sensor_id(?s)∧ little(?s,?rh)∧ swrl:greaterThan(?rh,"2")⇒MopupNeeds(?s, little) | If the mop up needs of deep organic material are greater than "2", then the support in fire is "little". |
| sensor_id(?s)∧moderate(?s,?rh)∧swrl:greaterThan(?rh,"10") ⇒MopupNeeds(?s, moderate) | If mop up needs of deep organic material are greater than "10", then supporting a fire is moderate. |
| sensor_id(?s)∧Difficult(?s,?rh)∧swrl:greaterThan(?rh,"20") ⇒MopupNeeds(?s, difficult) | If mop up needs of deep organic material are greater than "20", then supporting a fire is "difficult". |
| sensor_id(?s)∧difficultandextended(?s,?rh)∧ swrl:greaterThan(?rh,"30") ⇒MopupNeeds(?s,difficultandExtented) | If mop up needs of deep organic material are greater than "30", then supporting a fire is "difficult and extended". |
| sensor_id(?s)∧Difficultandextensive(?s,?rh)∧ swrl:greaterThan(?rh,"40")⇒ MopupNeeds(?s,difficultandextensive) | If mop up needs of deep organic material are greater than "40", then supporting a fire is "difficult and extensive". |

**Build Up Index (BUI)** - is a numerical evaluation of the full amount of gasoline available for a start that connects DMC and DC. A rating over 40 shows that fuel levels become high open for a start. A rating over 60 shows that the utilization of fuel can be extraordinarily high and fire control will be risky. Ranges belongs 0 to unlimited but it sub divided into five groups. First group 1 to 15, second group 16 to 30, third group 31 to 45, fourth group 46 to 59 and fifth group 60 to unlimited. According to this ranges "Difficulty of Control" are observe ( i.e easy, not difficult , very difficult etc.). Based on this ranges the Inference rules are made for give restriction on the ontology which shown in Table 4.

Table 4. **Build Up Index swrl rules**

| Rules Created for BUI | Explanation |
|---|---|
| sensor_id(?s)∧easy(?s,?rh)∧swrl:greaterThan(?rh,"1") ⇒DifficultyofControle(?s, easy) | If the difficulty of control of a gasoline is greater than "1", then it is "easy" to control a fire. |
| sensor_id(?s)∧notdifficult(?s,?rh)∧swrl:greaterThan (?rh,"16") ⇒DifficultyofControle(?s,notDifficult) | If the difficulty of control of a gasoline grater is greater than "16", then "not difficult" to control fire. |
| sensor_id(?s)∧difficult(?s,?rh)∧swrl:greaterThan (?rh,"31")⇒DifficultyofControle(?s, difficult) | It is "difficult" to control a fire if the difficulty of control of a gasoline grater is greater than "31." |
| sensor_id(?s)∧verydifficult(?s,?rh)∧swrl:greaterThan (?rh,"46")⇒DifficultyofControle(?s,veryDifficul) | If the difficulty of controlling a gasoline grater is greater than "46," then controlling a fire is "very difficult." |
| sensor_id(?s)∧extremelydifficult(?s,?rh)∧swrl:greaterThan (?rh,"60")⇒DifficultyofControle(?s, extremelyDifficult) | Controlling a fire is "extremely difficult" if the difficulty of controlling a gasoline grater is greater than "60." |

**Initial Spread Index(ISI)** - is a numerical assessment of the fire spread usual musicality not long after the start Consolidate wind speed and FFMC impacts in the pace of inciting without the impact of the variable Fuel Amount a rating of 10 shows a high expansion rate and a rating in any occasion 16 shows an incredibly fast expansion rate. Ranges belongs 0 to unlimited but it sub divided into five groups. First group 1 to 3, second group 4 to 7, third group 8 to 12, fourth group 13 to 15 and fifth group 16 to unlimited. According to this ranges "Rate of Spread" are observe ( i.e slow, fast, moderately fast etc.). Based on this ranges the Inference rules are made for give restriction on the ontology which shown in Table 5.

Table 5. **SWRL rules for Initial Spread Index**

| Rules Created for ISI | Explanation |
|---|---|
| sensor_id(?s)∧slow(?s,?rh)∧swrl:greaterThan(?rh,"1") ⇒RateofSpread(?s,slow) | If the rate of spread of fire is greater than "1" then "slow" spread of fire. |
| sensor_id(?s)∧moderatelyfast(?s,?rh)∧swrl:greaterThan(?rh, "4") ⇒RateofSpread(?s, moderatelyFast) | If the rate of spread of fire is greater than "4", then "moderately fast" spread of fire. |
| sensor_id(?s)∧fast(?s,?rh)∧swrl:greaterThan(?rh,"8") ⇒ RateofSpread(?s, fast) | If the rate of spread of fire is greater than "8", then "fast" spread of fire. |
| sensor_id(?s)∧veryfast(?s,?rh)∧swrl:greaterThan (?rh, "13") ⇒ RateofSpread(?s, very_fast) | If the rate of fire spread of fire is greater than "13", then it will be a "very fast" spread of fire. |
| sensor_id(?s)∧extremelydifficult(?s,?rh)∧swrl:greaterThan (?rh,"16")⇒RateofSpread(?s,extremelyDifficult) | If the rate of spread of fire is greater than "16", then it will be "extremely difficult" to control the spread of fire. |

**Fire Weather Index (FWI)** - is a numerical firepower rating that combines ISI and BUI. The FWI shows the sensible intensity of a fire and is sensible as a general record of fire chance. A FWI of more than 30 is considered remarkable. Ranges belongs 0 to unlimited but it sub divided into five groups. First group 1 to 5, second group 6 to 12, third group 13 to 20, fourth group 21 to 29 and fifth group 30 to unlimited. According to this ranges "Fire Intensity" are observe ( i.e low, high, very high etc.). Based on this ranges the Inference rules are made for give restriction on the ontology which shown in Table 6.

Table 6. **SWRL rules for Fire Weather Index**

| Rules Created for FWI | Explanation |
|---|---|
| sensor_id(?s)∧low(?s,?rh)∧swrl:greaterThan(?rh,"1") ⇒ FireIntensity(?s,low) | If the fire intensity is greater than "1 ", then "low" spread of fire. |
| sensor_id(?s)∧moderate(?s,?rh)∧swrl:greaterThan(?rh,"6") ⇒ FireIntensity(?s, moderate) | If the fire intensity is greater than "6 ", then "moderate" spread of fire. |
| sensor_id(?s)∧high(?s,?rh)∧swrl:greaterThan(?rh,"13") ⇒FireIntensity(?s,high) | If the fire intensity is greater than "1 ", then "high" spread of fire. |
| sensor_id(?s)∧veryhigh(?s,?rh)∧swrl:greaterThan(?rh,"21") ⇒FireIntensity(?s,:veryhigh) | If the fire intensity is greater than "21", then "very high" spread of fire. |
| sensor_id(?s)∧extreme(?s,?rh)∧swrlb:greaterThan(?rh,"30") ⇒FireIntensity(?s, extreme) | If the fire intensity is greater than "30 ", then the "extreme" spread of fire is very difficult to control. |

Table 7. **SWRL rules for Rain and Wind Speed**

| Rules Created for Rain | Explanation |
|---|---|
| sensor_id(?s)∧Rain(?s,?rh)∧swrlb:greaterThan(?rh,"1") ⇒startRaining(?s,FireStop) | If rain is greater than "1", when it starts raining, there is no chance of fire being spread by "fire stop". |
| **Rules Created for Wind Speed** | **Explanation** |
| sensor_id(?s)∧windspeed(?s,?rh)∧swrlb:greaterThan(?rh,"50") ⇒WindSpeed(?s,veryhigh) | If it crosses the parameters that are given and there is no rain, then it will be very difficult to control the fire. According to the range above, "50" indicates the "very high" risk level. |

## VIII. IMPLEMENTATION

*1) **The Proposed Ontology for Observational Processes**:* The ontology is built via seven step process.

1) Determine the ontology's domain and scope.
2) Think about repurposing existing ontologies.
3) A list of significant concepts in ontology (core classes).
4) Classify the classes and their hierarchies.
5) Slots are used to define the properties of classes.
6) he facets of the slots must be defined.
7) Instances are created.

In our model, Collect many concepts (entities) that are related to our domain of interest. Then it categorize those concepts in nouns and verbs. Noun forms the basis for classes, and verbs form the basis of object properties and occurrences. Then we put those concepts in our Ontology at their appropriate position. Afterwards define data properties and object properties for those concepts. At last make Inference rules for fire detection by relating those concepts using SWRL.

Fig. 5 depicts the entire user interface and how users interact with the forest fire management system using Sparql queries. It also shows how the data source can be mapped into the Resource Description Framework and then presented in graphical format using Onto Graph(also called Knowledge Graph[4]. Inference rules are used to give restrictions on the knowledge graph. To obtain a result, use a Sparql query.

*2) **The Core module**:* **Core Classes:-**

**SSN:Observation** - In this class we added dc, ffmc, dmc, isi, bui and fwi. For covering all the values and ranges of these elements.

**SSN:Sensor** - In this class we added sensor id as a subclass, with sensor id we added sensor number with location.

**SSN:Property** - In this class we added type as a subclass of, with type we added rain, relative humidity, temperature and wind.
**SSN:SensorOutput** - Covers all values which are produced by sensors and also covers which type of value it is.
**SSN:Feature Of Interest** - Covers all the dmc, ffmc, bui, which help to make fire weather index.
**SSN:UnitOfMeasure** - In this class we add all units which define (i.e millimeter, celsius, kilometer, etc).

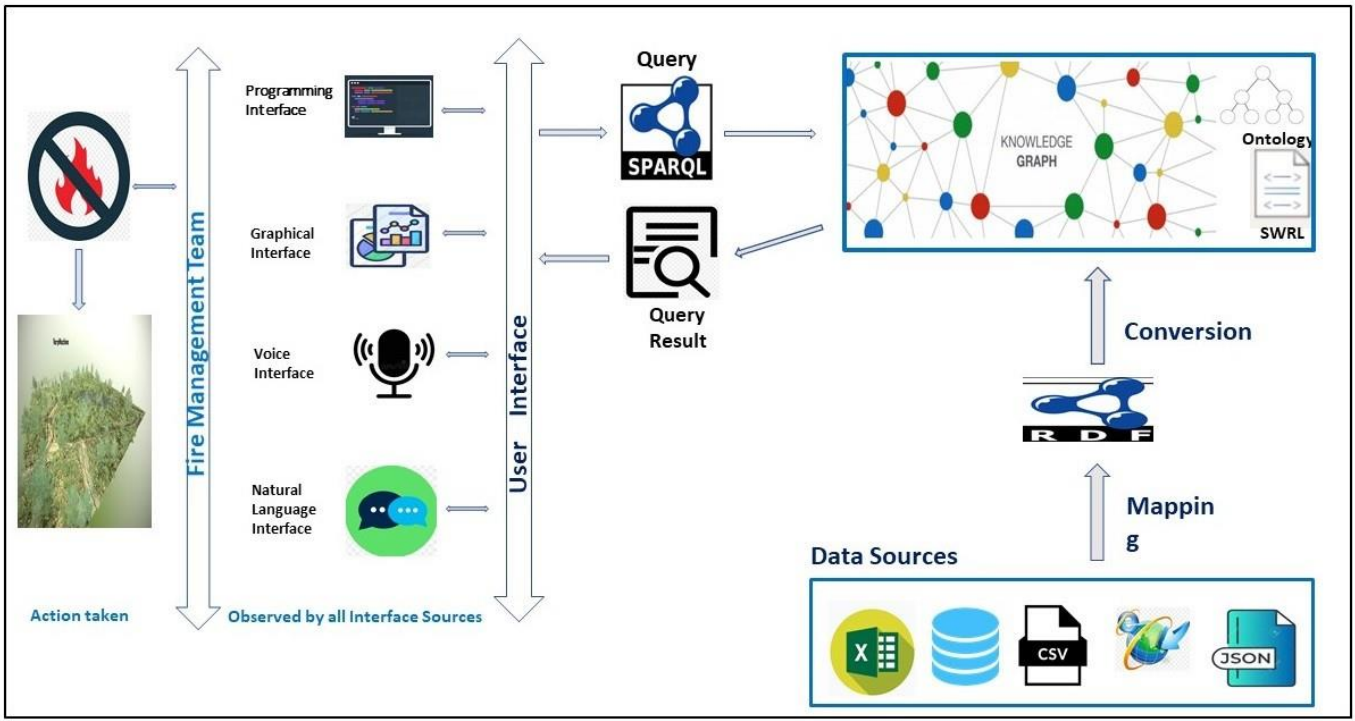

Fig. 5. **User Interface for Forest Fire Management system**

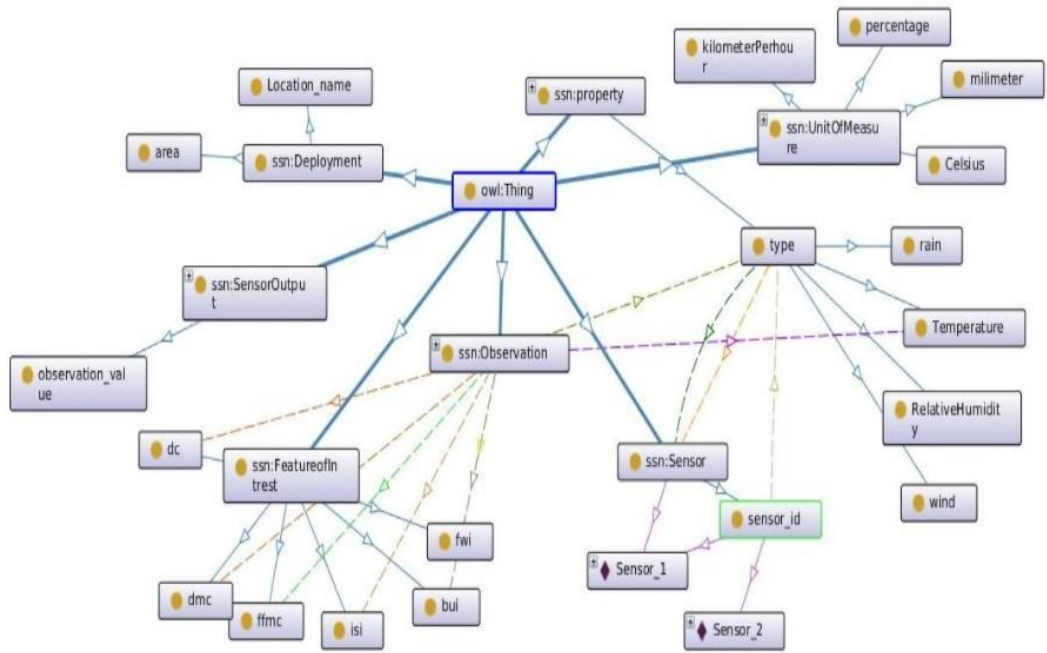

Fig. 6. **Instance of SSN ontology using Onto Graph [30]**

**SSN:Property** - In this class we added a type of subclass in this subclass all types of properties (i.e rain, relative humidity etc.)

*3) Data properties:* in a class definition describe attributes of instances.They are used for establishing relations of attributes with class. While defining data properties we consider domain and range for the data (concept). Domain represents an entity for which the data is defined and range represents the type of data. Range can be of any type viz. boolean,integer, string etc. Each data property is the sub-data-property of owl:topDataProperty, as shown in Fig. 7

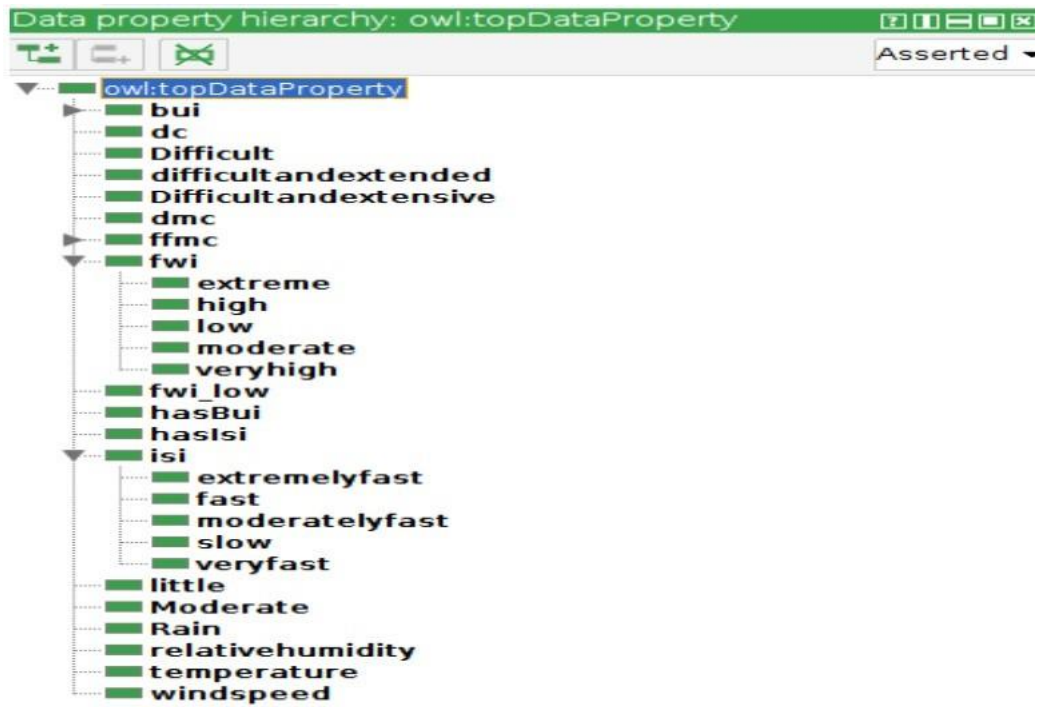

Fig. 7. **Framework Data property Protege Tool [32]**

The *ssn:WindSpeed* tell about wind speed rate.
The *ssn:Temperature* comes under dataproperty which tells about temperature according to the locations.
The *ssn:relativehumidity* tells about different humidity.
The *ssn:rain* tells that if it rains there is no chance of fire activity.

*4) Object Properties:* Used for establishing relationships between two entities. The domain is a property that links a class description to a property. It asserts that the subjects of property statements belong to the class extension indicated by the class description. A range saying declares that the estimations of this property must have a place with the class augmentation of the class depiction or to information esteems in the predefined information goal that is shown in Fig.8
The *DifficultyofControle* is an object property which represents the level of control of fire.
The *fireduetohuman* this class represents the fire that happens due to human like due to nuclear plant, chemical factory etc.
The *fireDuetoNature* this is used to represent the fire that happens due to nature like strom, friction of trees etc.
The *FireIntensity* this class tells about the speed of fire expansion level.
The *IgnitionPotential* tells about the level of harm.
The *MopupNeed* explains about the importance of moderate duff levels and medium-sized woody particles being consumed

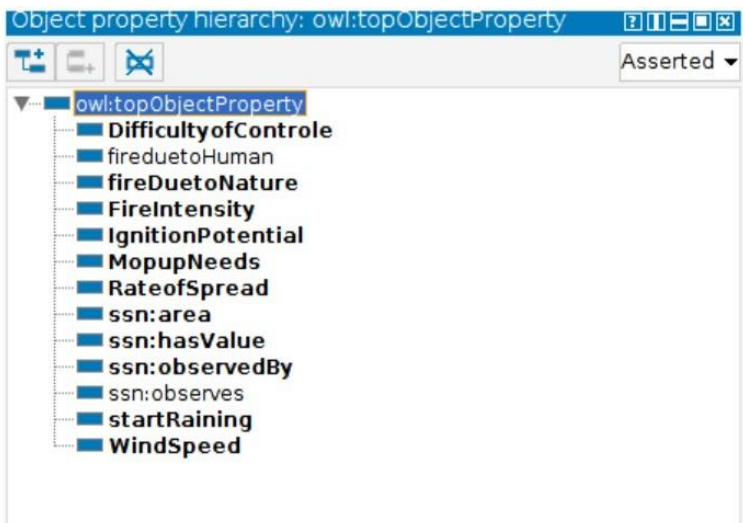

Fig. 8. **Framework Object property using Protege tool [31]**

by fire.

The *RateofSpread* tells about how fast fire spreads in the limit of time.

The *ssn:area* represents the sensor area where it was fixed.

The *ssn:hasValue* represents the sensor output value.

The *ssn:observedBy* denoted as an observation result which is observed by any specific sensor number.

The *ssn:observes* represents no specific output and observation can be denoted by this object property.

The *startRaining* this object property represents rain.

The *WindSpeed* represents wind speed about any location.

*5) **Observation by Sensor 3:** Fig. 9 show the output of the sensor- 3. After running the Pellet Reasoner (version 2)[5] which is Plugin in Protege (version 5.5.0)[6], if there is any inconsistency in ontology, then it gives a warning. Otherwise, it shows the output.

Observation of Sensor 3 as per Fig.9

In Step 1. (Note: Everything marked in yellow in Protégé was inferred by the reasoner based on the data and the ontology's axioms.) The "potential ignition" is "difficulty" comes under the range of (0- 74) as per Table 2. The "rate of spread" is "slow" under range 0-3 (as per Table 5), but due to "rain", there is no chance of fire.

In Step 2. Sensor 3 belongs to "sensor_id" class.

In Step 3. three Data property are used "Difficult", "slow" and "Rain".

In Step 4. three Infrence rules are used.

In Step 5. three Object property are used "IgnitionPotential", "startRaing" and "RateofSpread".

Rule 1 *sensor_id(?s) ∧ Difficult(?s,?rh) ∧ swrl:greaterThan(?rh,"1") ⟹ IgnitionPotential(?s, difficult)*

The moisture material of surface litter and other repaired fine fills is given a numerical rating called ignition potential. It demonstrates the fine energizes' overall simplicity of initiation and combustibility. It gives the Difficulty level of ignition potential in that area (as per follow Fig.10).

Rule 2 *sensor_id(?s) ∧ slow(?s,?rh) ∧ swrl:greaterThan(?rh,"1") ⟹ RateofSpread(?s,slow)*

Rate of Spread is a numerical evaluation of the ordinary musicality of The fire spread not long after the start Consolidate wind and FFMC impacts in the pace of inciting without the impact of the variable fuel amount(as per follow Figure 9).

Rule 3 *sensor_id(?s) ∧ Rain(?s,?rh) ∧ swrlb:greaterThan(?rh,"2") ⟹ startRaining(?s,Fire _Stop)*

Starting raining never possible spread of fire(as per follow Fig.10).

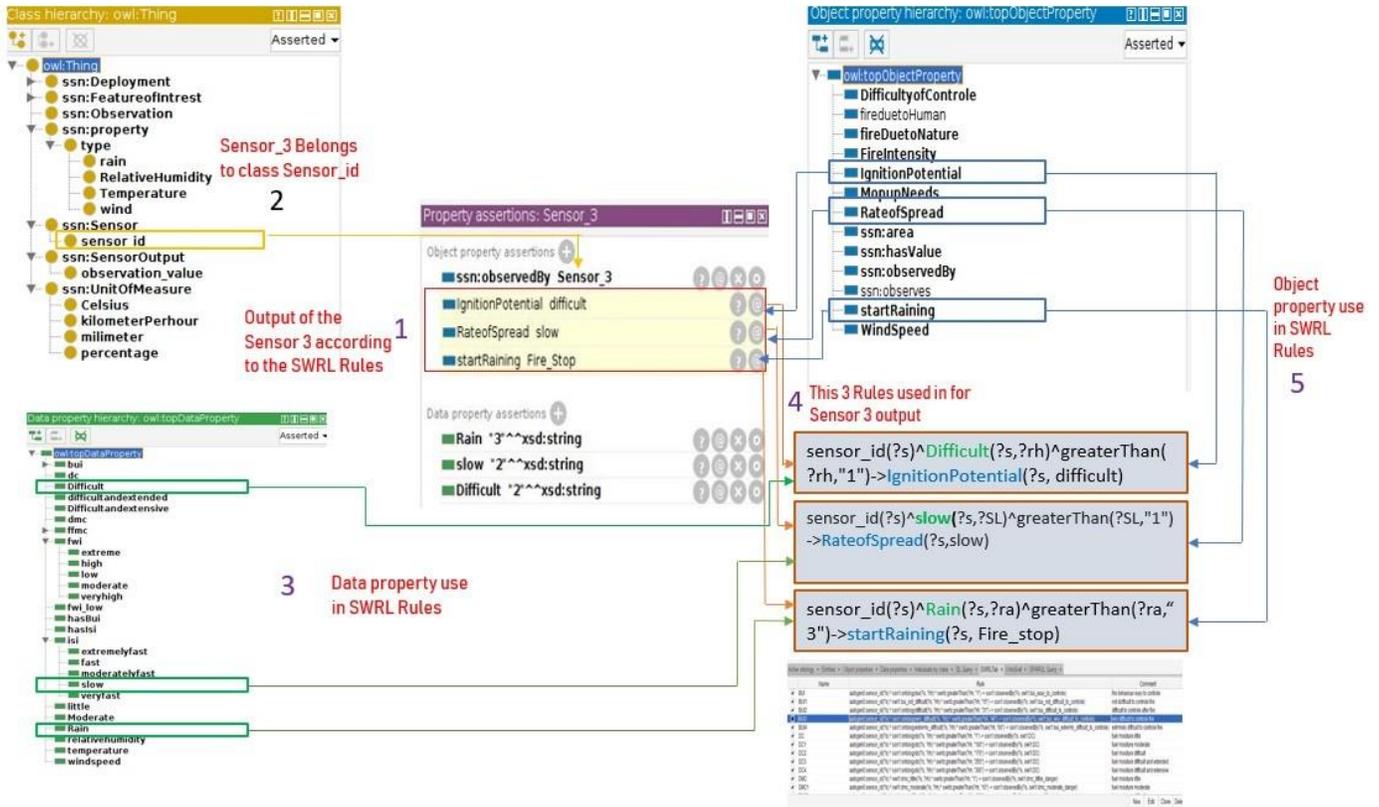

Fig. 9. **Framework of background working process Protege tool for sensor 3**

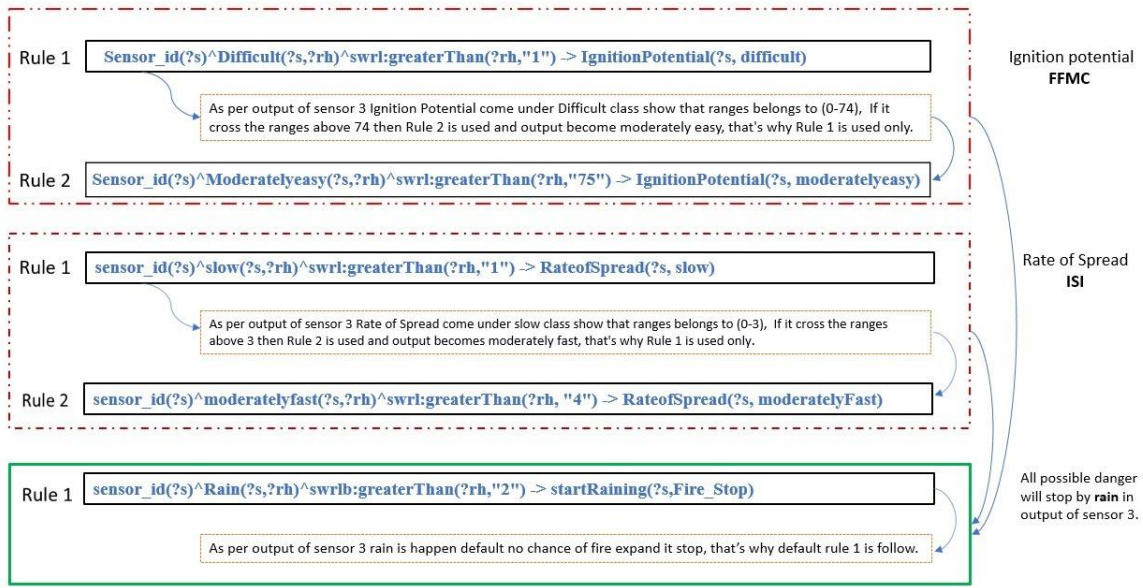

Fig. 10. **Swrl rules the working process for sensor 3 output.**

*6) Observation by Sensor 2:* Observations of sensor 2 are from different locations here the output given by protege after running the reasoner. There is a fire in this location, and the fire intensity is moderate, so fire management takes further decisions. Output covers( Difficulty of control shown in Fig.11, fire intensity shown in Fig. 12 and MoupNeeds shown in Fig.13).

**Difficulty of control** - It is a numerical evaluation of the full amount of gasoline available for a start that connects DMC and DC. A grade of more than 40 indicates that fuel levels are high when the vehicle starts. A rating over 60 shows that the utilization of fuel is extraordinarily high and fire control is be risky. Which is shown in Table 4.

Rule used in Observation output of sensor 2 for *Difficulty of control*.
sensor_id(?s)∧notdifficult(?s,?rh)∧swrl:greaterThan(?rh,"16")⇒DifficultyofControle(?s,notDifficult)

Fig. 11. **Fire Weather Index Difficulty of Control and which rule follows for this.**

**Fire intensity** - is a numerical rating of fire power that joins ISI and BUI. The FWI shows the sensible intensity of a fire and is sensible as a general record of fire chance. A FWI of more than 30 is considered remarkable. Which is shown in Table 6.

Rule used in Observation output of sensor 3 *Fire intensity*.
sensor_id(?s)∧moderate(?s, ?rh) ∧ swrl:greaterThan(?rh, "6")⇒ FireIntensity(?s, moderate)

Fig. 12. **Fire Weather Index Fire Intensity and which rule follows for this.**

**Mop up Needs** -The Mop up Needs code shows the significance of the substance's texture to fire. This rating is usually used to describe the substance's sogginess.Which is shown in Table 3.

Rule used in Observation output of sensor 3 *Moup needs.*

sensor_id(?s)∧moderate(?s,?rh)∧swrl:greaterThan(?rh,"10")⇒MopupNeeds(?s, moderate)

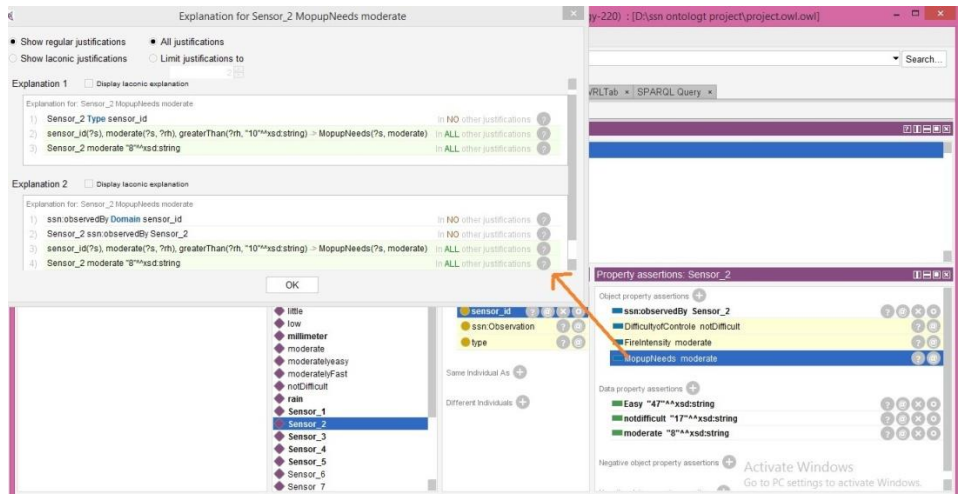

Fig. 13. **Fire Weather Index mop up needs and which rule follows for this.**

*7)* ***Deployment of the Proposed Approach****:* To implement the semantic-enhanced Discovery System, the suggested solution employs the SSNO and rules. It takes data from an observational process and maps it to the SSNO so that rules-based inference can be performed.

**Use of a Case Scenario for Demonstration**

The forest fire management system is a web-based platform that enables users to perform various tasks related to the management of forests. It consists of various modules that can be used to collect and process data. When a threshold value is reached, the system will automatically inform about the event's details and provide warning content. It will also evaluate the rules for the system. Figure 14 displays the results of a query for a monitoring station that illustrate the outcomes of the fire alert [21] .

**Comparison with other ontologies**

Table 8. How SSN Ontology is better for other Ontology

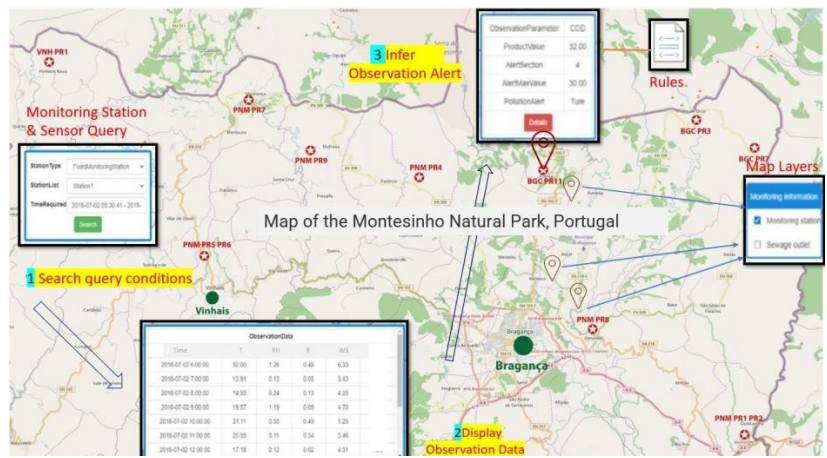

Fig. 14. **The discovery service consists of a collection of data collected by the software system, which includes monitoring stations, observation products, and warning systems.**

| Features | Semantic Sensor Network Ontology (SSNO)) | Observational Process Ontology (OPO) |
|---|---|---|
| Target | Ontology development for describing sensors | a linguistics description of the empirical method for higher discovery |
| Important ideas | Sensors, system, actuation's, and observation are all words that come to mind while thinking about sensors | Sensors, data from observations, and analysis of observations, and observation result are all words that come to mind when thinking about observation |
| A wide range of topics | Comprehensive | Broad in scope |
| Adoption | There are a variety of instances, such as weather stations | Water quality monitoring and pollution |
| Best features | Interoperability and broader applicability | It aids in the explanation of observational procedures |
| Weakest features | Sufficient observational process descriptions, especially for FWI | Insufficient description of sensor metadata |

## IX. Result

*1) **SPARQL Query Result**:* SPARQL [32] is an RDF Query Language. RDF is a labelled, directed graph data format used to describe information on the Web. This specification defines the syntax and semantics of the SPARQL query language for RDF. SPARQL may be used to define searches across a wide range of data sources, independent of whether the data is stored natively as RDF or is seen as RDF through middleware.SPARQL may query both required and optional graph patterns, as well as their conjunctions and disjunctions. SPARQL also allows for extendable value testing and query constraints based on the underlying RDF graph. SPARQL queries can produce results sets or RDF graphs. Fig.15 is frame view of SPARQL Query tab which Plugin in Protege tool. It show query result the wind speed of all sensor id. The output of this query is in URI [34] formats. According to the query result we decide the rate of spread of fire in different locations [25] according to the sensor id.

Table 9. Sparql Query for all sensor id with filter wind speed greater than 40. This query run by Protege which shown in Fig. 15.

```
PREFIX rdf: <http://www.w3.org/1999/02/22-rdf-syntax-ns#>
PREFIX owl: <http://www.w3.org/2002/07/owl#>
PREFIX rdfs: <http://www.w3.org/2000/01/rdf-schema#>
PREFIX xsd: <http://www.w3.org/2001/XMLSchema#>
SELECT ?Sensor_id ?WindSpeed
WHERE { ?Sensor_id ?observedBy ?WindSpeed
FILTER (?WindSpeed >40.00) }
```

Fig. 15. **Frame view of SPARQL Query tab which show the wind speed of all sensor.**

Table 10.  Sparql Query for all sensor id with filter rain more than 1. This query run by Protege which shown in Fig. 16
In Fig. 16 the query results show that rain grater than 1 of all sensor id. According to this result it decide that if raining is going on then no chance of fire in that location.

Fig. 16. **Frame view of SPARQL Query tab which show the rain of all sensor.**

Same way all query will executed with different scenario, this is handle by Fire Management team.

In Fig.17 graph of 78 days, which are not common regular days, includes those days with the most dramatic weather change data in Montensinho Natural Park. Which shows how FFMC, DMC, and DC are related to each other. Whenever DC can change together, DMC also changes, which was clear through Fig.17.
In Different condition how this model work it clear through example.

```
PREFIX rdf: <http://www.w3.org/1999/02/22-rdf-syntax-ns#>
PREFIX owl: <http://www.w3.org/2002/07/owl#>
PREFIX rdfs: <http://www.w3.org/2000/01/rdf-schema#>
PREFIX xsd: <http://www.w3.org/2001/XMLSchema#>
SELECT ?Sensor_id ?startRAIN
WHERE { ?Sensor_id ?observedBy ?startRAIN
FILTER (?startRAIN >1.00) }
```

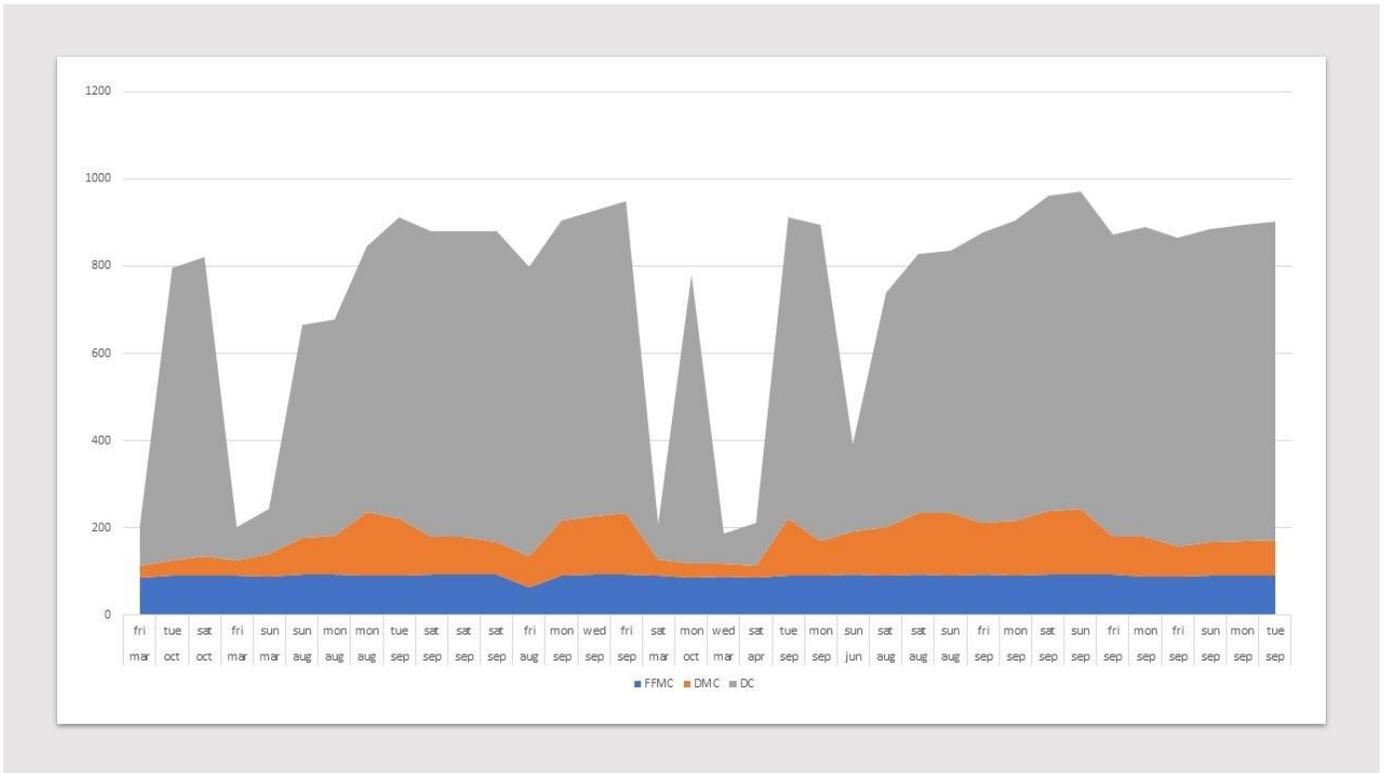

Fig. 17. **Fuel Moisture Code graph**

**Example** FFMC= 95, DMC= 47, and DC= 321 These numerical values indicate that:
•Fine fuels are exceedingly easy to ignite.
• All types of fuel will be included in the fire.
• Extreme fire behaviour is almost certain.
• Fires on the Spot.
The fire behaviour indices (ISI, BUI, and FWI) predict the initial spread, total fuel availability, and fire intensity.

**Example** FFMC= 88, DMC= 27, and DC= 122
These numeric values indicate that:
• Fine fuels are easy to ignite.
• The fine fuels, as well as the medium and duff layer fuels, will be involved in the fire.
• There will be no deep seated fire.

**Example** if we take the value of ISI= 6, BUI=115 and FWI=23
These numeric values indicate:
• Initial spread should be slow to moderate.
• There is a large amount of fuel accessible for combustion.
• High-intensity fires are a possibility.
A hot, but slow-moving fire in general. After a long, dry time, this is the type of fire that is most likely to erupt on a windless day in mid-summer.

## X. CONCLUSION

An ontology for forest fire detection that is employed in forest fire management systems was proposed in this paper. This ontology was created via a semantic network process as part of the Semantic Sensor Networks Incubator Group. This study proposes semantically enhanced fire detection alert inference methods for integrating wind speed, temperature, rain, and relative humidity sensors, as well as deep knowledge mining from observational data. The forest fire detection model was able to be used for the protection of wild fires using the rules. The proposed rules introduce a conceptual framework for the query results to be integrated into a data observation system collected for the purpose of making a DSS for forest fire management. The system uses Montesinho's natural park data specifications to describe and illustrate various pollution and fire safety alert details. This article recommends using a semantic mix of observation techniques to solve the problem of protection against forest fires in the Natural Park of Montesinho using SSN Ontology.

It was also discovered that the system's running time rises while processing big ontologies owing to the enormous amount of information, indicating that the framework's scalability has to be improved. In this regard, it considered adopting parallel processing techniques, which allow us to use several processors to examine various portions of the ontologies concurrently, improving the execution time.

In the future, merging different existing ontologies for more conditions will cover more conditions, describe more rules and instances will be employed to gather observational data and extract more meaningful information. For reasoning, the system must still be able to consult external semantic sensor web services.

## ABBREVIATION

| | |
|---|---|
| FFMC | Fine Fuel Moisture Code |
| DMC | Duff Moisture Code |
| DC | Drought Code |
| ISI | Initial Spread Index |
| BUI | Buildup Index |
| FWI | Fire Weather Index |
| SSNO | Semantic Sensor Network Ontology |
| SWRL | Semantic web rule language |
| DSS | Decision Support System |
| RDF | Resource Description Framework |